\documentclass[letterpaper,twocolumn,fleqn]{article} 

\usepackage{ist}
\usepackage{graphicx}
\usepackage{subfig}
\usepackage{placeins}
\usepackage{dblfloatfix}
\usepackage[font=small,labelfont=bf]{caption}

\title{Rehabilitating the ColorChecker Dataset for\\ Illuminant Estimation}

\author{Ghalia Hemrit$^1$, Graham D. Finlayson$^1$,  Arjan Gijsenij$^2$, Peter Gehler$^3$, Simone Bianco$^4$, Brian Funt$^5$, Mark Drew$^5$ and Lilong Shi$^6$\\
$^1$School of Computing Sciences, University of East Anglia; Norwich, United Kingdom\\
$^2$AkzoNobel; Amsterdam, The Netherlands\\
$^3$Amazon; T\"ubingen, Germany\\
$^4$Imaging and Vision Laboratory, University of Milan-Bicocca; Milan, Italy\\
$^5$School of Computing Science, Simon Fraser University; Vancouver, Canada\\
$^6$Samsung Semiconductor Inc; Pasadena, USA\vspace{-2.5\baselineskip}}

\date{} 

\begin{document} 

\maketitle 

\thispagestyle{empty} 

\begin{abstract}
In a previous work, it was shown that there is a curious problem  with the
benchmark ColorChecker dataset for illuminant estimation. To wit, this dataset
has at least 3 different sets of ground-truths. Typically, for a single
algorithm a single ground-truth is used. But then different algorithms, whose
performance is measured with respect to different ground-truths, are compared
against each other and then ranked. This makes no sense. We show in this paper that there are also errors in how each
ground-truth set was calculated.  As a result, all performance rankings based on
the ColorChecker dataset -- and there are scores of these -- are inaccurate.

In this paper, we re-generate a new `recommended' ground-truth set based on the
calculation methodology described by Shi and Funt. We then review the
performance evaluation of a range of illuminant estimation algorithms. Compared
with the legacy ground-truths, we find that the difference in how algorithms
perform can be large, with many local rankings of algorithms being reversed. 

Finally, we draw the readers attention to our new `open' data repository which, we hope, will allow the ColorChecker set to be rehabilitated and once again become a useful benchmark for illuminant estimation algorithms.

\end{abstract}

\section{1.Introduction}
\label{sec:intro}
\begin{figure}[htb]
\centering
\subfloat{\includegraphics[width=0.49\columnwidth]{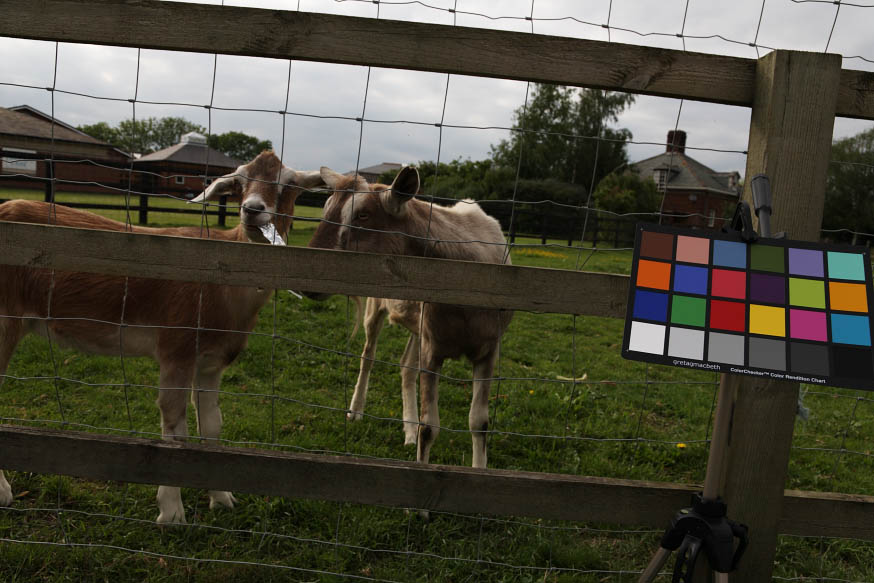} } \hfill
\subfloat{\includegraphics[width=0.49\columnwidth]{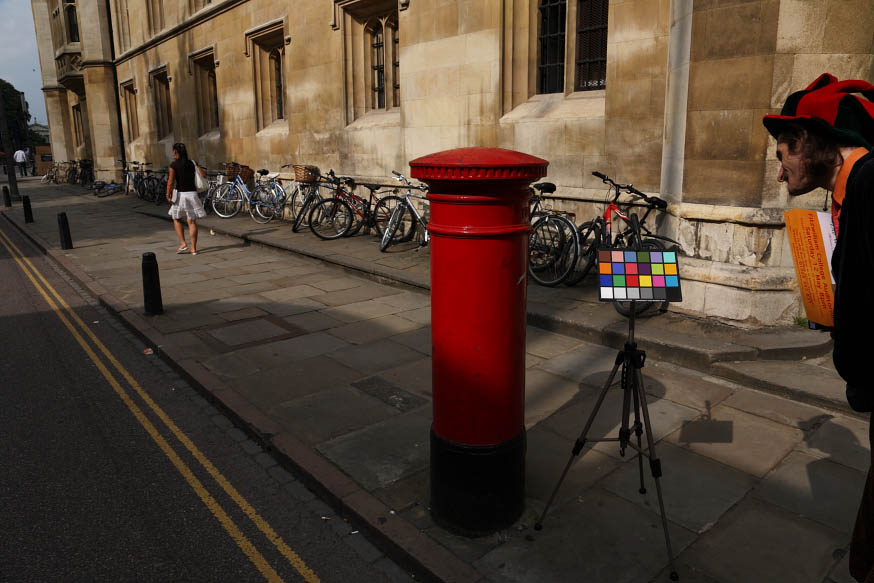} } \\
\vspace*{-2.8mm}
\subfloat{\includegraphics[width=0.49\columnwidth]{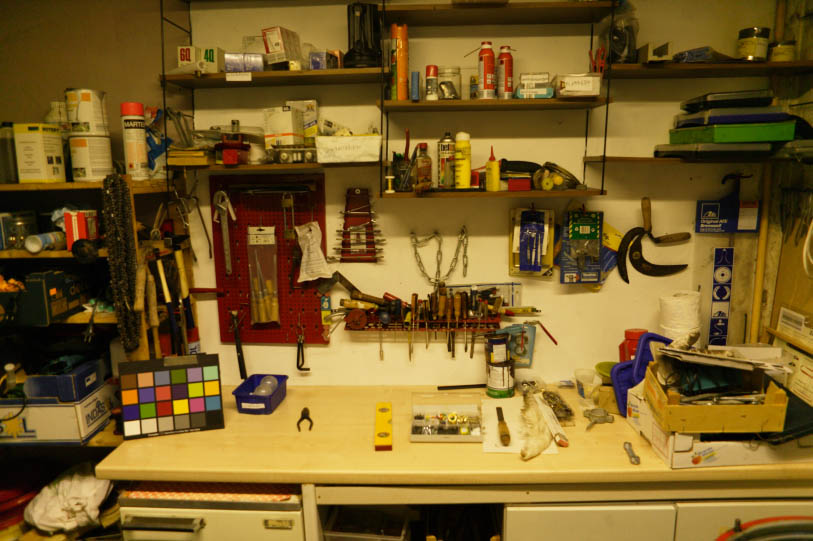} }\hfill
\subfloat{\includegraphics[width=0.49\columnwidth]{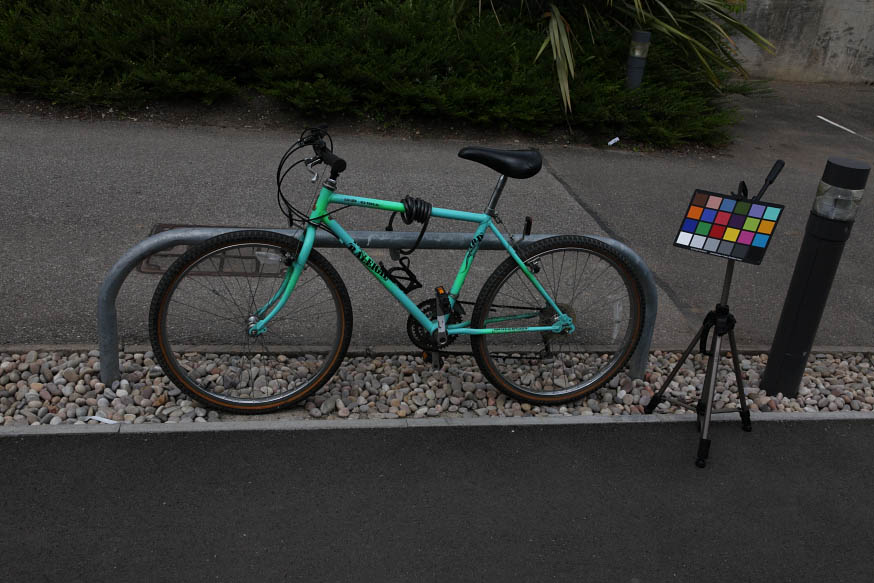} }\vspace*{-2.8mm}
\subfloat{\includegraphics[width=0.49\columnwidth]{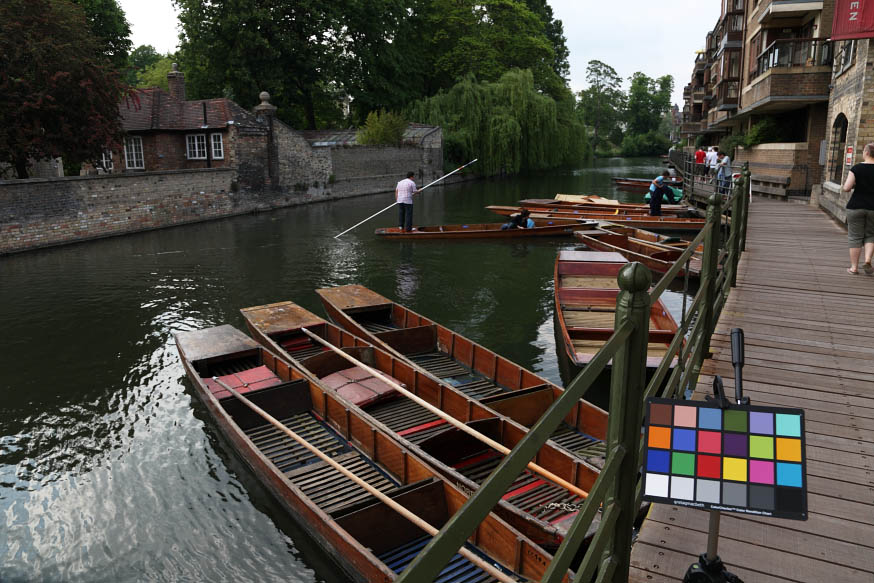} }\hfill
\subfloat{\includegraphics[width=0.49\columnwidth]{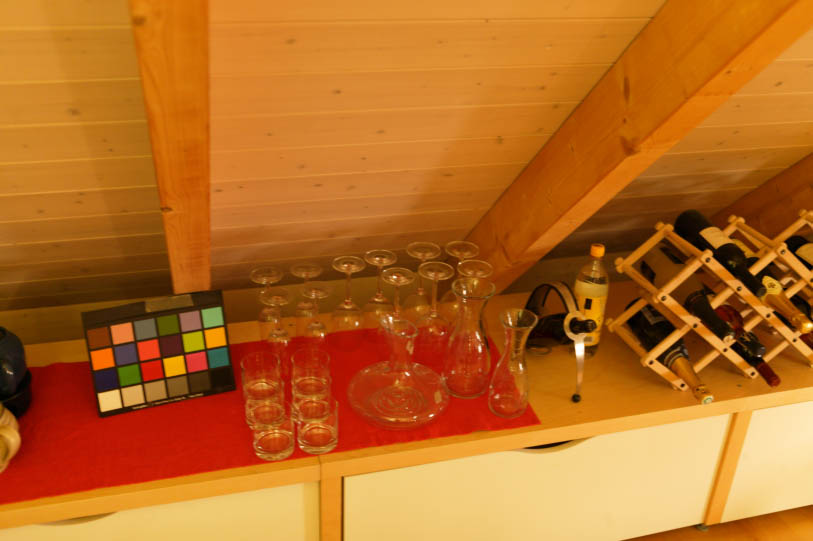} }
\vspace*{1mm}
  \caption{Images from the Macbeth ColorChecker dataset; here the images are the camera pipeline outputs.}
  \label{fig:example}
\end{figure}

The ColorChecker dataset was introduced by Gehler et~al.\ in 2008
\cite{Gehler2008}. It has 568 images of various daily and ordinary tourist
scenes (see Figure~\ref{fig:example}) mainly taken in Cambridge with two popular
cameras, the Canon 1D and the Canon 5D. The ColorChecker dataset is probably the most widely used dataset in evaluating the performance of algorithms for illuminant estimation.

The goal of most illuminant estimation algorithms is to infer the chromaticity of the light. The reader will notice that each image in
Figure~\ref{fig:example} has a Macbeth ColorChecker placed in the scene. The RGB
from the brightest non-saturated achromatic patch (see last row of color chart)
is the correct answer -- or ground-truth -- for illuminant
estimation, according to the calculation methodology described by Shi and
Funt \cite{datasetshi}. In fact, most of the recent work in illuminant
estimation uses a linear version of the dataset, with linear images reprocessed by Shi and Funt \cite{datasetshi} from Gehler's original raw images \cite{Gehler2008}. 

Naturally, the performance of illuminant estimation algorithms -- which adopt a
range of strategies to infer the RGB of the illuminant -- is determined by how
well they predict the ground-truth illuminant color. Because we cannot
distinguish between a bright scene dimly lit and the converse (because the
intensity of the illuminant is not recovered), the measure which is used to
quantify the accuracy of the estimation is the angular error. Given a set of
angular errors, various statistical summaries are used to summarise performance.
These include the mean, median or 95\% quantile angular error. Given a set of
algorithms and their summary performance statistics, it is natural to rank all
the algorithms (according to the statistic), and then to conclude that algorithm
A is better than B which is better than C (e.g.\ if their respective means are in an ascending order).

While there are many image sets (with respect to which, algorithms might be evaluated in
performance and ranked), the ColorChecker dataset is the most widely used.
Indeed, perhaps the most comprehensive survey of algorithms performance was
carried out by  Gijsenij et~al.\ \cite{Gijsenij2011}. The results reported there are often quoted in more recent papers with the paper now cited  $\sim$400 times (April 13, 2018). The algorithms considered in this survey also form the basis of the color constancy evaluation site (colorconstancy.com).  

Of course, this evaluation process makes perfect sense, {\it in principle}. In
practice, we observed in \cite{Finlayson2017} a surprising flaw in the
methodology. Namely, we discovered that there are 3 different sets of
ground-truths, one ground-truth -- {\bf SFU} -- on
www.cs.sfu.ca/$\scriptstyle\mathtt{\sim}$colour/data/shi\_{}gehler/ \cite{datasetshi} and two ground-truths -- {\bf Gt1} and {\bf Gt2} -- on
colorconstancy.com. Unfortunately, these 3 ground-truths are used in a haphazard
way. Specifically, an author will adopt one of the three ground-truths, then
will evaluate the algorithm and, say, calculate the median angular error. In the
next step, this median is compared against other algorithms' median errors. But, the medians for these competitor algorithms have been calculated with respect to {\bf all 3}  ground-truth correct answers. This makes no sense, especially since \cite{Finlayson2017} demonstrated that using different sets of ground-truths would drastically affect the ranking of algorithms. At the time of writing this paper, there is no definitive ranking of illuminant estimation algorithms (for the ColorChecker dataset).

Here, we seek to explain why we find ourselves in this multiple ground-truth
world. We then go on to make a new `recommended' ground-truth ({\bf REC}) for
the community. Then, we present the results evaluating algorithms using this recommended ground-truth and compare our results to the benchmark evaluation and rankings. The rankings relative to the new recommended ground-truth reveal for the first time the actual pecking order in illuminant estimation for the ColorChecker dataset.
 
In Section 2, we discuss how we compute the new {\bf REC} ground-truth
and explain why this ground-truth differs from the other 3 in the literature. In
section 3, we present some analysis of the relative performance of different
algorithms using {\bf REC} compared to the legacy data. The paper ends with a short conclusion.
 
\section{2. Derivation of the RECommended Ground-truth}

We adopt directly the methodology introduced by Shi and Funt \cite{datasetshi} to re-process the raw images and re-calculate the ground-truth set of illuminants of the ColorChecker dataset. However, we have used our own code and in the interests of transparency, we will  make our code accessible online \cite{dataset}. 

\begin{figure}[ht]
\begin{center}
\includegraphics[width=1\columnwidth]{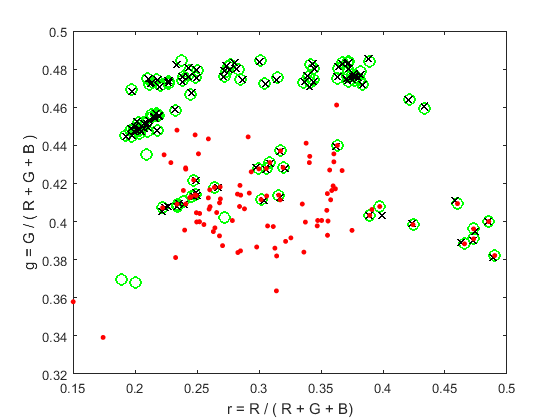}
\end{center}
  \caption{REC ground-truth chromaticities are plotted as black crosses. The green circles and red dots  respectively denote the SFU and Gt1 ground-truths.}
  \label{fig:gamut}
\end{figure}

Readers who `click through' to this code and the data will find some additional
information. Specifically, we explain that our data repository is `open'. We set
forth a mechanism for the community to provide further suggested modifications
to this ground-truth dataset (if necessary). In the short term, we will monitor
any suggestions that the community makes and incorporate these suggestions -- if
they are significant -- into the data. This would lead to a new recommended ground-truth. This said, we have been quite careful in our calculations so it is our hope that our new {\bf REC} ground-truth will stand the test of time. If a modification is suggested, then both the current {\bf REC} and the updated version will be retained and labelled with clear time stamps.

\begin{figure}[ht]
\begin{center}
\includegraphics[width=1\columnwidth]{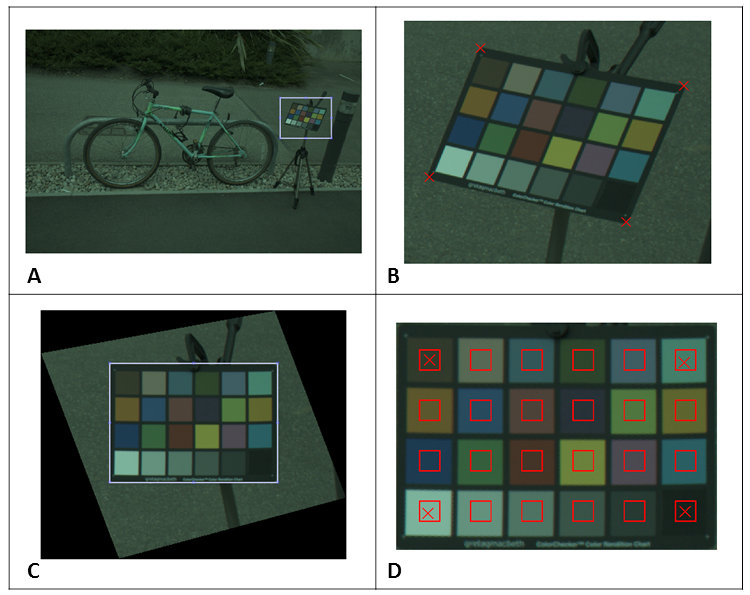}
\end{center}
  \caption{The 4 main steps of the color chart processing. A) we select roughly the chart area in the image. B) we select the 4 corners of the chart. C) the chart image is geometrically transformed to be in front of the camera, then we select more precisely its contour. D) we select the centers of the 4 corners patches as well as one square of interest in one patch, this square selection is then automatically replicated over all patches. The medians per channel of the achromatic red-square regions are calculated.}
  \label{fig:processing}
\end{figure}

Let us outline the {\bf REC} ground-truth calculation. We reprocessed Gehler's
raw images to obtain linear demosaiced images using {\em dcraw} \cite{dcraw}. Every image has a ColorChecker in the scene. This color chart provides a reference for measuring the ground-truth illuminant. In each image, the ColorChecker chart is selected and the median RGB from the brightest achromatic patch (ranked by average of the selected squares with no digital count$>$3300 [each image is in 12 bits]) defines a ground-truth illuminant. The main steps of the color chart processing are presented in Figure~\ref{fig:processing}. 
 
One point to highlight is the importance of using the same patch or patches when defining the R, G and B of the light color. We found that this property was not properly enforced in the calculation of the 
{\bf SFU} ground-truth. In fact, we found 3 images where saturated color channels/patches were not 
correctly identified which resulted in having the ground-truth R, G and B values not taken from the 
same patch, in these three cases. 

An important detail about the ColorChecker dataset is the `black level'. This is zero for Canon 1D images but is 129 for the Canon 5D. One important contribution of the Shi-Funt calculation methodology is the subtraction of the camera black level from the ground-truth. The offsets were estimated from the minimum pixels across the whole dataset \cite{datasetshi}. 

\nocite{Barron2017, vandeweijer2007, Gijsenij2010, vandeweijer2007, Gehler2008, Rosenberg2003, Bianco2015, Joze2012, Gijsenij2010, Gijsenij2010, Schmid2007, Schmid2007, Gijsenij2011e, Chakrabarti2012, Schmid2007, Bianco2010, Bianco2015, Bianco2010, vandeweijer2007, Land1971, Finlayson2004, Xiong2004, Buchsbaum1980, Nishino2003} 

\begin{table}[!b]
    \caption{Ranking of  23 algorithms in terms of median recovery error for REC vs SFU vs Gt1; the Minkowski norm p and the smoothing value $\sigma$ are the optimal parameters.} 
    \label{tab:tab1}
    \begin{center}
    \includegraphics[width=0.45\textwidth]{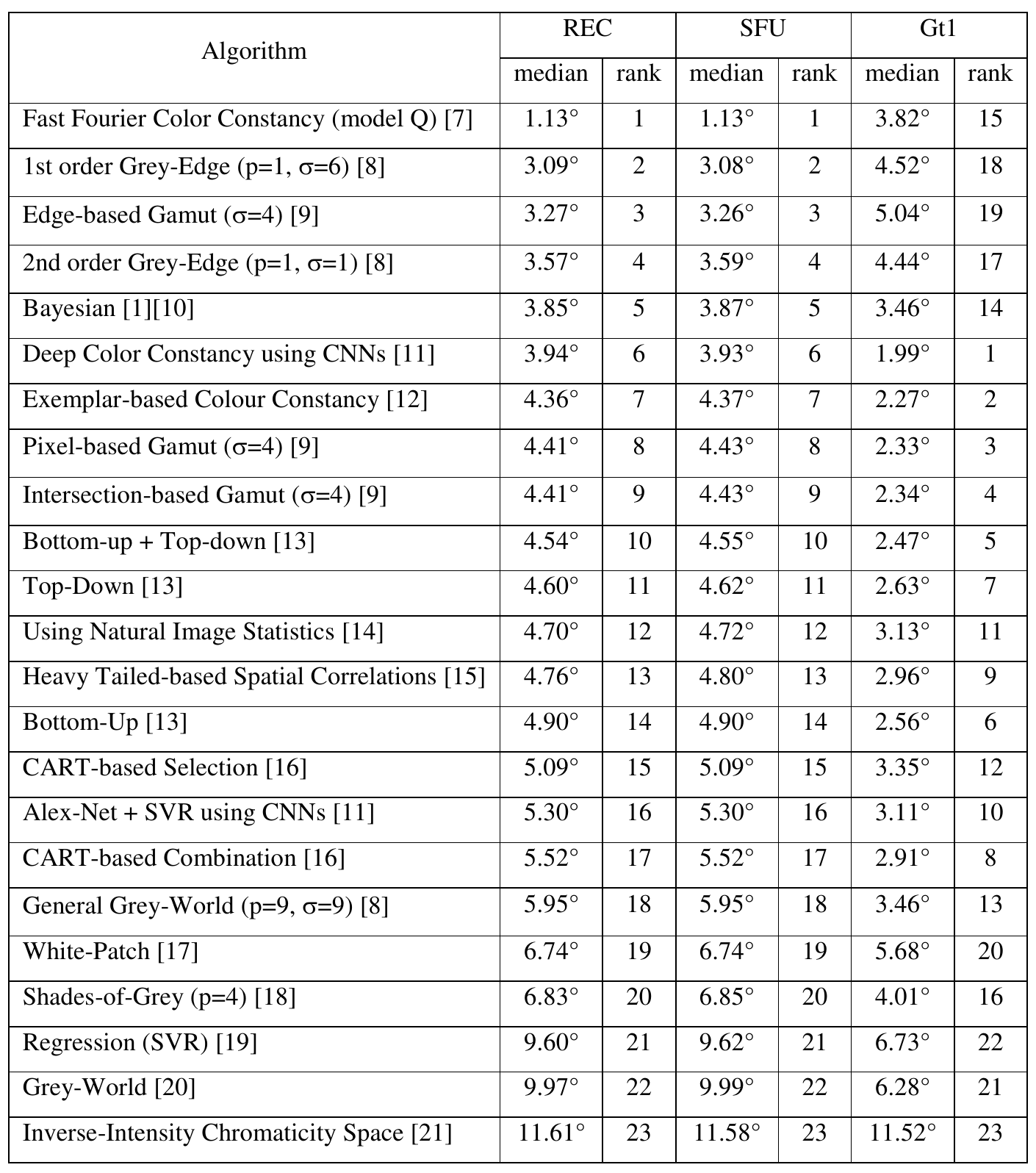}
    \end{center} 
\end{table}

Let us discuss the ground-truths. The ground-truth set we call {\bf SFU} has the
568 estimates of the RGB lights colors and it appears currently on the SFU site.
On colorconstancy.com, there are two additional ground-truths, {\bf Gt1} and
{\bf Gt2}. {\bf Gt1} was calculated and put
online by Gijsenij  in 2011. It refers to the same data as {\bf SFU} but
without the black level being subtracted. {\bf Gt1} was calculated based
on the color chart patches RGB data which is available on the SFU site
\cite{datasetshi} but without subtracting the offset. A little investigation on
our part led us to the following explanation: on the {\it current} site
\cite{datasetshi}, it is stated that the black level needs to be subtracted
before using the data: ``Note that for most applications . . . the black level
offset will still need to be subtracted from the original images". When looking
at the same web-page using the waybackmachine.org
site -- that allows recovering web-pages from many years ago -- the previous
statement is not present and instead we can read "[the processing] also takes
into account that the Canon 5D has a black level of 129, which we subtract." We
spoke to a number of researchers in the field and asked them to read the current and past web-pages. Everyone agreed
that on reading the past instructions they could
easily have made the same `mistake' (and not subtracted the black level).
Gijsenij believes the web-page instructions are a plausible explanation of why he did not subtract the black point.

Researchers who have been in the field a
long time have been using {\bf Gt1} but more recent workers are using {\bf SFU}. Potentially, all of these researchers {\it think} that they are using the same ground-truth. But in fact, the data are very different.

The third legacy ground-truth dataset we call {\bf Gt2}. It is very
similar to {\bf Gt1} (and so, also, different from {\bf SFU}). The reason for
the difference between {\bf Gt1} and {\bf Gt2} is explained by Bianco
\cite{BiancoVC}: ``we noticed that for some images the Macbeth ColorChecker
coordinates (both the bounding box and the corners of each patch) were wrong and
thus the illuminant ground-truth was wrong." In Figure~\ref{fig:gamut}, for a
subset of the 568 images, we compare the chromaticity distributions of our new
{\bf REC} ground-truth, {\bf SFU} and {\bf Gt1} ({\bf Gt2} is not shown as it is
almost the same as {\bf Gt1}). The reader will notice that the {\it current}
{\bf SFU} ground-truth is close to our newly calculated {\bf REC}ommended
ground-truth except for a few points. These look to be set apart from the rest
of the data i.e.\ they appear to be outliers, in some sense (note that, in Figure~\ref{fig:gamut}, the 3 green circles to the left of the plot that do not overlap with the black crosses). These are data points in {\bf SFU} that are different from {\bf REC}. 

We posit that the outliers are due to the problems in calculating correctly the
bounding boxes (Bianco's observation) and to our own discovery that the white
point was on occasion drawn from different achromatic patches (for R versus G
versus B). Despite this, the preponderance of the data is in, more or less, precise alignment. In contrast, the points in {\bf Gt1} are far from {\bf REC}. 

\begin{table}[!h]
    \caption{Ranking of 23 algorithms in terms of median reproduction error
    \cite{Finlayson2016} for REC vs SFU vs Gt1; the Minkowski norm p and the smoothing value $\sigma$ are the optimal parameters.}
    \label{tab:tab2}
    \begin{center}
    \includegraphics[width=0.45\textwidth]{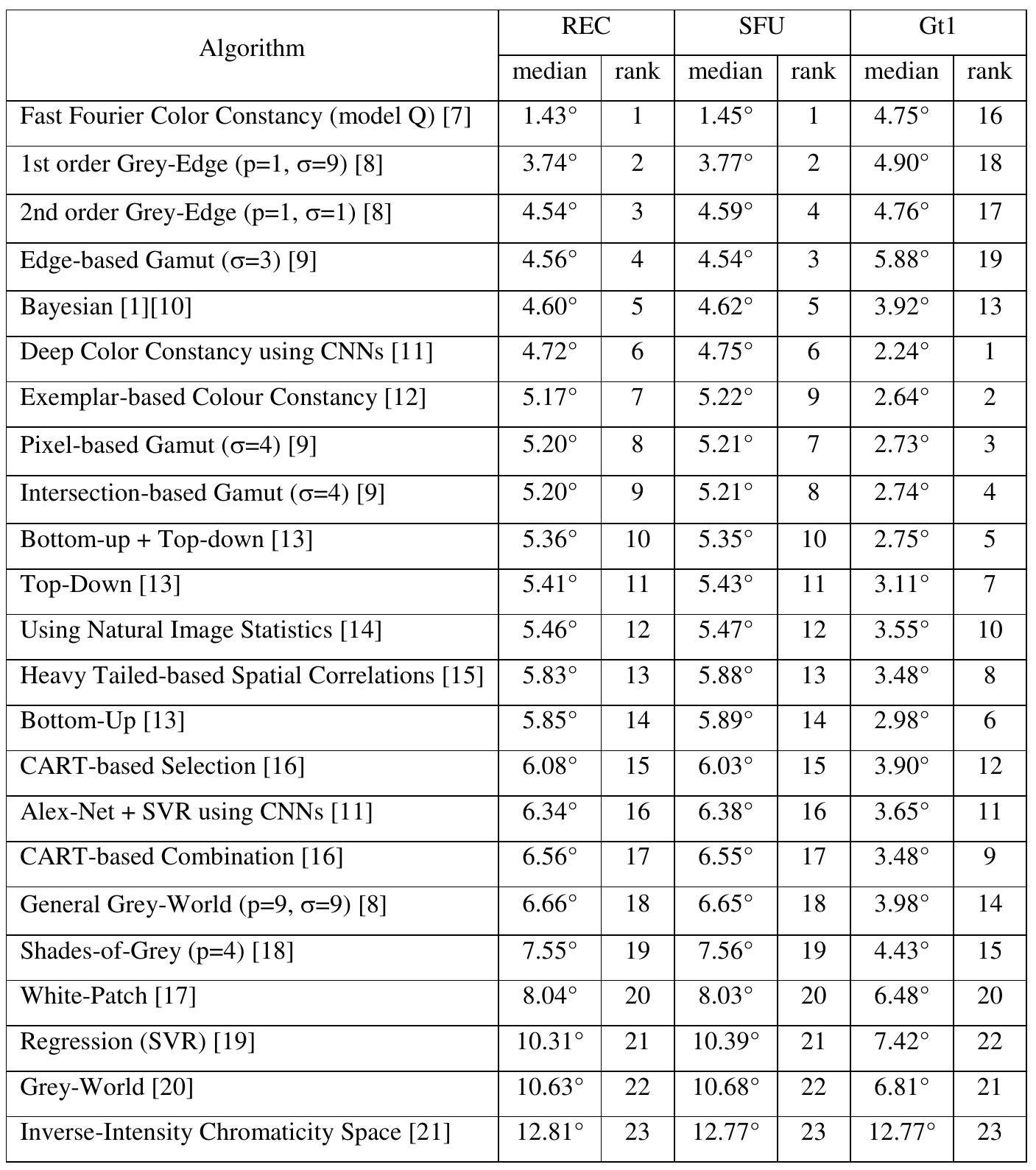}
    \end{center}
\end{table}
\vspace{-0.3in}

\section{3. Evaluating the Performance of Illuminant Estimation Algorithms}

In what follows, we provide performance evaluation and ranking of 23
illuminant estimations for the {\bf REC}ommended and the legacy {\bf SFU} and
{\bf Gt1} ground-truths. In Table~\ref{tab:tab1}, the 23 algorithms are ranked
according to the median recovery angular error (i.e.\ the conventional error
measure, which gives the angle between the estimated illuminant RGB
vector and the ground-truth illuminant RGB). The ranking is the same for {\bf
REC} and {\bf SFU}, although the median errors are slightly different. This is
expected as (recall Figure~\ref{fig:gamut}) the number of differences between
{\bf REC} and {\bf SFU} is small (due to errors in the calculation). However,
the ranking of algorithms according to the {\bf Gt1} dataset is markedly
different (due as opposed to a different methodology of calculation).

The five best algorithms with {\bf Gt1} are not in the top 5 according to {\bf
REC}. Vice versa, the 5 best algorithms according to {\bf REC} are among the
worst-performing algorithms with {\bf Gt1}. The algoithms Edge-based Gamut
Mapping \cite{Gijsenij2010}, 2nd order Grey-Edge \cite{vandeweijer2007} and
Bayesian \cite{Gehler2008}\cite{Rosenberg2003} are, for example, in reverse
order. Fast Fourier Color Constancy \cite{Barron2017} -- which is in significant
part built on top of a machine learning algorithm -- is best according to the
{\bf REC}ommended ground-truth but is the 15th best on {\bf Gt1}. This may not
be surprising as this algorithm was trained on {\bf SFU}. Deep Color Constancy
using CNNs \cite{Bianco2015} is 6th based on {\bf REC} and top ranked based on
{\bf Gt1}. Again, this is not surprising since this algorithm was trained on {\bf Gt2} (which, we recall, is very similar to {\bf Gt1} but very different from {\bf SFU} and {\bf REC}).

In Table~\ref{tab:tab2}, we consider the ranking of the 23 illuminant estimation
algorithms in terms of median reproduction angular error \cite{Finlayson2016}.
Reproduction error is an angle-type error that evaluates how well a white surface is reproduced. Since image reproduction is the goal of most illuminant estimation algorithms, reproduction error provides a more useful measure of algorithm performance. This time, results with {\bf REC} and {\bf SFU} locally differ. Once again, the results with {\bf Gt1} are significantly different. Finally notice that the ranks for the same ground-truth but recovery vs reproduction angular error results in different rankings.

In Tables~\ref{tab:tab1} and ~\ref{tab:tab2}, we do not include the results for {\bf Gt2} because they are very comparable to {\bf Gt1}, however we invite the reader to consult our previous work on this topic \cite{Finlayson2017}. A more complete survey is accessible on colorconstancy.com.

\FloatBarrier
\section{Conclusion}
Illuminant estimation algorithms have been evaluated and compared on the
benchmark ColorChecker dataset with at least three different ground-truths, with one
of the three being very different to the other two. In addition, we found that
all three of these sets of ground-truths were inaccurately or
incorrectly calculated in the sense that small errors were made. The problem of multiple ground-truths and calculation errors has led to misleading results in the performance evaluation of illuminant estimation algorithms. 

In this paper we have introduced a new {\bf REC}ommended ground-truth for this
dataset which we hope rehabilitates the ColorChecker dataset. Broadly, we
followed the methodology set forth by Shi and Funt but using our own code and we
corrected a few errors made (e.g.\ those reported in \cite{BiancoVC}). We
re-evaluated all the algorithms on the widely used comparison site for
illuminant estimation algorithms, colorconstancy.com. We invite the community to refer to what we hope is a more definitive comparison in future research.

\section{Acknowledgments} 
This research was supported by EPSRC Grant M001768 and  Apple Inc. 

\small
\bibliographystyle{ieeetr}
\bibliography{CIC2018}

\end{document}